\documentclass[lettersize,journal]{IEEEtran}
\usepackage{amsmath,amssymb,amsfonts}

\usepackage{mathtools}
\usepackage{algorithm}
\usepackage{algpseudocode}
\usepackage{array}
\usepackage[caption=false,font=normalsize,labelfont=sf,textfont=sf]{subfig}
\usepackage{textcomp}
\usepackage{stfloats}
\usepackage{csquotes}
\usepackage{xcolor,soul}
\usepackage[dvipsnames]{xcolor}

\usepackage{adjustbox}
\usepackage{etoolbox}
\usepackage{hyperref}
\usepackage{multirow}
\usepackage{url}
\usepackage{verbatim}
\usepackage{graphicx}
\usepackage{subcaption}
\usepackage{subfig}
\usepackage{makecell}
\usepackage{gensymb}
\usepackage{multirow}
\usepackage{enumitem}
\usepackage{diagbox}
\usepackage{todonotes}
\usepackage{parskip}
\usepackage{orcidlink}
\usepackage{float}
\usepackage{enumitem}
\usepackage{slashbox}
\def\BibTeX{{\rm B\kern-.05em{\sc i\kern-.025em b}\kern-.08em
    T\kern-.1667em\lower.7ex\hbox{E}\kern-.125emX}}
\usepackage{balance}
\usepackage[numbers,sort&compress]{natbib}
\usepackage{etoolbox}

\begin{document}
\title{Information-Aided DVL Calibration}

\author{\IEEEauthorblockN{Zeev Yampolsky\IEEEauthorrefmark{1},\orcidlink{0009-0003-9122-7576}, and Itzik Klein, \orcidlink{0000-0001-7846-0654}}
\IEEEauthorblockA{\\The Hatter Department of Marine Technologies\\
Charney School of Marine Sciences, University of Haifa,
Haifa, Israel}

\thanks{\IEEEauthorrefmark{1}Corresponding author: Zeev Yampolsky, zyampols@campus.haifa.ac.il}}

\maketitle

\begin{abstract}
    The Doppler velocity log (DVL) velocity measurements are critical to the accuracy of autonomous underwater vehicle (AUV) navigation solutions and, consequently, to mission success. To ensure accurate measurements, the DVL is commonly calibrated before mission start while the AUV sails on the water surface, receiving global navigation satellite system (GNSS) signals that provide accurate reference measurements. Conventionally, Kalman filter-based approaches are employed during calibration to estimate the scale factor and misalignment errors. However, in certain environments, GNSS signals may be unavailable, rendering conventional calibration impossible and forcing the use of uncalibrated DVL measurements, which degrades navigation performance. To address this limitation, this work proposes information-aided calibration (IAC) with two main contributions: first, improving the accuracy of conventional Kalman filter-based calibration in GNSS-enabled environments, and second, enabling GNSS-free DVL self-calibration. Using real-world AUV datasets, the proposed IAC models achieve up to a $20\%$ average improvement in GNSS-enabled environments and up to a $35\%$ improvement in velocity vector estimation during GNSS-free DVL self-calibration. Overall, the proposed approach improves navigation accuracy, reduces navigation drift, and consequently enhances mission reliability.
\end{abstract}


\section{Introduction}\label{sec:intro_sec}
Autonomous underwater vehicles (AUVs) are widely utilized in marine and oceanic exploration, research, and industrial applications. Their applications range from marine biological surveys and Arctic ice-cap exploration to marine geoscience and autonomous infrastructure inspection \cite{feng2023automatic,williams2012monitoring,barker2020scientific,wynn2014autonomous,ridao2010visual}. Due to the autonomous nature of AUV operations, accurate navigation solutions are critical for mission success and the vehicle's safe return \cite{liu2025positioning,miao2026physics,sheng2025ekf}. Commonly, underwater navigation relies on the fusion of a strapdown inertial navigation system (SINS) and a Doppler velocity log (DVL) \cite{wang2019novel,cohen2024seamless,luo2023line,qian2023ins}, where the DVL provides accurate velocity measurements processed in a nonlinear estimation framework, such as the extended Kalman filter (EKF) \cite{ribeiro2004kalman,cohen2025adaptive} or unscented Kalman filter (UKF) \cite{wan2000unscented,levy2026adaptive,diker2026neural}, to correct and bound inertial drift.\\
The DVL provides velocity measurements by transmitting acoustic beams toward the seabed and estimating each beam’s velocity using the Doppler shift effect upon receiving the reflected signals, thereby providing direct velocity measurements of the AUV \cite{braginsky2020correction,pan2024novel}. Nevertheless, DVL measurements still contain error sources such as misalignment between the DVL and AUV body frames, scale factor errors, and biases. Therefore, to ensure accurate velocity measurements, the DVL must be calibrated prior to mission start to estimate the deterministic, time-invariant error terms \cite{ning2023research,wang2021quasi,li2022calibration}. Traditionally, calibration is performed while the AUV sails on the water surface to receive global navigation satellite system (GNSS) signals, which serve as reference measurements for DVL calibration \cite{liu2022gnss,li2025novel,xu2020novel,xu2022novel,wang2022online}. During the calibration process, the AUV is required to perform predefined maneuvers to increase the observability of the estimated states, commonly the scale factor and misalignment errors, within the employed Kalman filter framework \cite{liu2022gnss,xu2022novel,huang2025gnss}.\\
Nevertheless, recent works have proposed alternative approaches for DVL calibration, including optimization-based methods \cite{zhang2026novel}, genetic algorithms \cite{wang2020model}, and data-driven frameworks \cite{yampolsky2025dcnet}, demonstrating promising results. Furthermore, researchers have shown that calibration can be performed using straight-line or nearly constant-velocity trajectories \cite{wang2022online,yampolsky2025dcnet}. However, two main challenges in DVL calibration remain unresolved: first, how can DVL calibration be performed in GNSS-denied environments? Second, can the accuracy of traditional Kalman filter-based DVL calibration be further improved using information-aiding approaches?\\
In this work, we address the aforementioned challenges by proposing information-aided calibration (IAC) in the form of velocity information aiding with two main objectives: first, to improve DVL calibration when GNSS is available, and second, to enable GNSS-free DVL self-calibration. The proposed approach is motivated by non-holonomic constraints, commonly utilized in land-based platforms such as vehicles \cite{engelsman2023information}. Such non-holonomic constraints assume zero velocity along specific body-frame axes and are incorporated as measurement updates to improve navigation accuracy \cite{groves2008principles}. However, to the best of our knowledge, limited work has investigated the incorporation of such concepts into AUV navigation \cite{yao2017imm}. Therefore, this work makes the following main contributions:
\begin{enumerate}
    \item \textbf{Information-Aided Calibration}: IAC framework for improved DVL calibration in both GNSS availability scenarios.
    \item \textbf{Zero Velocity information aiding}: In GNSS-enabled scenarios, we propose to employ IAC by utilizing zero velocity information aiding during the DVL calibration for improved accuracy and reduced calibration time.
    \item \textbf{GNSS-free IAC DVL self-calibration}: We offer a relaxed error model combined with a novel GNSS-free IAC velocity assumption to enable GNSS-free DVL self-calibration in GNSS-denied environments, where the alternative is the use of uncalibrated velocity measurements.
\end{enumerate}
We propose five modified IAC-based Kalman filter models. Two models are designed for GNSS-enabled environments and employ the same four-state formulation as their baseline counterparts. By incorporating the proposed IAC approach, these models achieve up to a $20\%$ improvement compared to their baselines. Three additional models are designed for GNSS-denied environments and employ a relaxed three-state formulation combined with the proposed GNSS-free IAC approach, enabling DVL self-calibration with an average improvement of $30\%$ and up to $35\%$ compared to the uncalibrated baseline. All proposed models were validated using real-world data collected by the University of Haifa AUV, demonstrating their robustness and applicability.\\
The rest of this paper is organized as follows: Section \ref{sec:prob_form} presents the DVL error model, and the Kalman filter framework. Section \ref{sec:prop_app_general} details the proposed IAC approaches for both GNSS availability scenarios. Section \ref{sec:res_gen} presents the real-world validation process and the results of the proposed approaches. Finally, Section \ref{sec:conc_sec} concludes the paper.
\section{Problem Formulation}\label{sec:prob_form}
This section presents the mathematical formulation of the DVL velocity measurements and the Kalman filter framework for the DVL calibration problem.
\subsection{DVL Velocity Error Model}\label{sec:dvl_vel}
Commonly, the following error model is applied during DVL calibration \cite{xu2022novel,liu2022gnss}:
\begin{equation}\label{eq:general}
    \tilde{\boldsymbol{v}}^d = (\mathbf{I}+\boldsymbol{\delta k})\mathbf{T}_{b}^{d}(\mathbf{T}_{n}^{b}\tilde{\boldsymbol{v}}^{n} + \boldsymbol{\omega}_{nb}^{b}\times \boldsymbol{l}) + \boldsymbol{\sigma}^{d}
\end{equation}
where $\tilde{\boldsymbol{v}}^d$ is the measured velocity expressed in the DVL frame, $\tilde{\boldsymbol{v}}^{n}$ is the measured velocity expressed in the navigation frame, $\mathbf{T}_{n}^{b}$ is the transformation matrix from the navigation frame to the body frame, $\mathbf{T}_{b}^{d}$ is the transformation matrix from the body frame to the DVL frame, $\boldsymbol{\delta k}$ is the scale factor, $\boldsymbol{\omega}_{nb}^{b}$ is the platform's angular velocity, $\boldsymbol{l}$ is the lever-arm between the platform's center of mass and the DVL, and $\boldsymbol{\sigma}^{d}$ is the zero-mean Gaussian white noise. 
Commonly, the term $\boldsymbol{\omega}_{nb}^{b}\times \boldsymbol{l}$ is neglected \cite{xu2020novel,zhang2026novel,wang2021quasi}. Therefore, the DVL calibration error model becomes \cite{wang2022online}:
\begin{equation}\label{eq:basic_calib_model}
    \tilde{\boldsymbol{v}}^{d} = ( \mathbf{I} + \boldsymbol{\delta k})\mathbf{T}_{b}^{d}\tilde{\boldsymbol{v}}^b + \boldsymbol{\sigma}^d
\end{equation}
The transformation matrix from the body frame to the DVL frame, $\mathbf{T}_{b}^d$, is measured during DVL installation by formulating the problem as Wahba’s problem \cite{wahba1965least} and solving it using model-based or data-driven approaches \cite{umeyama2002least,eggert1997estimating,damari2026resalignnet}. Nevertheless, small misalignment errors are still assumed to exist. Therefore, a more realistic transformation formulation that incorporates both the measured frame transformation and residual small misalignment errors is employed \cite{titterton2004strapdown}:
\begin{equation}\label{eq:euler_error_to_msalignment}
    \mathbf{T}_{b}^{d} = (\mathbf{I} + \boldsymbol{\phi}\times)\hat{\mathbf{T}_{b}^{d}}
\end{equation}
where $\hat{\mathbf{T}_{b}^{d}}$ is the estimated transformation, $\mathbf{T}_{b}^{d}$ is the true transformation, and $\boldsymbol{\phi} = \begin{bmatrix}\phi_x & \phi_y & \phi_z \end{bmatrix}^T$ represents the remaining small misalignment errors along the X, Y, and Z axes. Substituting \eqref{eq:euler_error_to_msalignment} into \eqref{eq:basic_calib_model} yields:
\begin{equation}\label{eq:final_dvl_error_model}
    \tilde{\boldsymbol{v}}^{d} = (\mathbf{I} + \boldsymbol{\delta k})(\mathbf{I} + (\boldsymbol{\phi}\times))\tilde{\boldsymbol{v}}_{GNSS}^d  + \boldsymbol{\sigma}^d
\end{equation}
where $\tilde{\boldsymbol{v}}^d_{GNSS}$ is the measured GNSS velocity expressed in the DVL frame: 
\begin{equation}\label{eq:gnss_in_DVL}
    \tilde{\boldsymbol{v}}_{GNSS}^d = \hat{\mathbf{T}}_{b}^{d}\tilde{\boldsymbol{v}}_{GNSS}^b
\end{equation}
\subsection{Kalman Filter Formulation}\label{sec:linear_KF_theory}
A linear Kalman filter system model is defined as \cite{bar2001estimation}:
\begin{equation}\label{eq:LKF_general}
    \dot{\boldsymbol{x}} = \mathbf{F}\boldsymbol{x} + \mathbf{W}
\end{equation}
where $\boldsymbol{x}$ is the state vector, $\dot{\boldsymbol{x}}$ is its time derivative, $\mathbf{F}$ is the continuous-time system matrix, and $\mathbf{W}$ is the system noise matrix. The measurement is defined as:
\begin{equation}\label{eq:z_cont_def}
    \boldsymbol{z} = \mathbf{H}\boldsymbol{x} + \boldsymbol{\sigma}_v 
\end{equation}
where $\mathbf{H}$ is the measurement matrix, and $\boldsymbol{\sigma}_v$ is the measurement noise, modeled as a zero-mean Gaussian white noise distribution such that $\boldsymbol{\sigma}_v \sim \mathcal{N}(0 , \boldsymbol{v}^2)$.\\
The discrete state vector time propagation is defined as follows \cite{groves2008principles}:
\begin{equation}\label{eq:x_minus_propogation}
    \hat{\boldsymbol{x}}^{-}_{k} = \mathbf{\Phi}_{k-1}\hat{\boldsymbol{x}}^{+}_{k-1}
\end{equation}
where $\hat{\boldsymbol{x}}^{-}_{k}$ is the apriori state estimate at the current time step $k$, $\hat{\boldsymbol{x}}^{+}_{k-1}$ is the posterior state estimate at the previous time step $k-1$, and $\mathbf{\Phi}_{k-1}$ is the discrete-time system matrix at time step $k-1$. The discrete-time error-state covariance propagation is given by:
\begin{equation}\label{eq:cov_prop}
    \mathbf{P}_{k}^{-} =  \mathbf{\Phi}_k \mathbf{P}_{k-1}^{+} \mathbf{\Phi}_k^T + \mathbf{Q}_{k}
\end{equation}
where $\mathbf{P}_{k}^{-}$ is the apriori state error covariance, $\mathbf{P}_{k-1}^{+}$ is the posterior state error covariance, and $\mathbf{Q}_{k}$ is the discrete process noise covariance. The discrete-time system matrix is obtained using a Taylor series expansion and neglecting high order terms \cite{paul2005fundamentals}:
\begin{equation}\label{eq:discrete_phi}
    \mathbf{\Phi}_k = \mathbf{I} + \mathbf{F}t 
\end{equation}
where $t$ is the system time interval. The process noise covariance is defined as $\mathbf{Q} = \mathrm{E}[\boldsymbol{w}\boldsymbol{w}^T]$. Accordingly, the discrete process noise covariance is given by:
\begin{equation}
    \mathbf{Q}_k = \int_{0}^{T_s}\mathbf{\Phi}(\tau)\mathbf{Q}\mathbf{\Phi}(\tau)^T dt
\end{equation}
Once a measurement becomes available, the Kalman gain is calculated as:
\begin{equation}
    \mathbf{K}_k = \mathbf{P}_{k}^{-} \mathbf{H}_k^T (\mathbf{H}_k \mathbf{P}_{k}^{-} \mathbf{H}_k^T + \mathbf{R}_{k})^{-1}
\end{equation}
where $\mathbf{R}_{k}$ is the discrete measurement noise covariance matrix at time step $k$, defined as $\mathbf{R}_{k} = \mathrm{E}[\boldsymbol{v}\boldsymbol{v}^T]$. The posterior state error covariance is computed as:
\begin{equation}
    \mathbf{P}_{k}^{+} = (\mathbf{I} - \mathbf{K}_k\mathbf{H}_k)\mathbf{P}_{k}^{-}
\end{equation}
Finally, the updated state vector is:
\begin{equation}
    \hat{\boldsymbol{x}}_{k}^{+} = \hat{\boldsymbol{x}}_{k}^{-} + \mathbf{K}_k(\boldsymbol{z}_{k} - \mathbf{H}_k\hat{\boldsymbol{x}}_{k}^{-})
\end{equation}
where $\hat{\boldsymbol{x}}_{k}^+$ is the state estimate at time step $k$ after the measurement update and $\boldsymbol{z}_k$ is the measurement at time step $k$.
\subsection{DVL Calibration}
The baseline state vectors for DVL calibration are defined as follows \cite{wang2022online}:
\begin{equation}\label{eq:model1_state_vec}
    \boldsymbol{x}_{M1} = \begin{bmatrix}
        \phi_x & \phi_y & \phi_z & \delta k
    \end{bmatrix}^T
\end{equation}
\begin{equation}\label{eq:model2_state_vec}
    \boldsymbol{x}_{M2} = \begin{bmatrix}
        \phi_x & \phi_y & \phi_z & \delta k_x
    \end{bmatrix}^T
\end{equation}
where $\boldsymbol{x}_{M1}$ is baseline model 1 (M1) state vector including three misalignment angles and a scalar scale factor which applies over all three velocity components. $\boldsymbol{x}_{M2}$ is the state vector of baseline model 2 (M2), consisting of the same three misalignment errors and a scale factor applied only to the X component. Since the estimated DVL error terms are constant and assumed time-invariant, the state derivatives of both baseline models are defined as zero:
\begin{equation}\label{eq:nulified_state}
    \dot{\boldsymbol{x}}_{M1} = \dot{\boldsymbol{x}}_{M2} = \boldsymbol{0}
\end{equation}
Subsequently, the system matrix is reduced to zero, $\mathbf{F} = \mathbf{0}\in \mathbb{R}^{4 \times 4}$. Thus, the discrete system matrix in \eqref{eq:discrete_phi} is:
\begin{equation}
    \mathbf{\Phi}_{k} = \mathbf{I} \in \mathbb{R}^{4\times4}
\end{equation}
Note that $\mathbf{\Phi}_{k}$ represents the system matrix for both baseline models, M1 and M2. \\
By expanding the parentheses and neglecting second order error terms, the DVL error model in \eqref{eq:final_dvl_error_model} is modified as follows:
\begin{equation}\label{eq:before_Z_and_H}
    \tilde{\boldsymbol{v}}^{d} = \tilde{\boldsymbol{v}}^d_{GNSS} + \boldsymbol{\delta k} \tilde{\boldsymbol{v}}^d_{GNSS} + (\boldsymbol{\phi} \times) \tilde{\boldsymbol{v}}^d_{GNSS}
\end{equation}
Subsequently, the following relationship between the DVL and GNSS measurements and the estimated error terms of M1 and M2 is obtained:
\begin{equation}
    \tilde{\boldsymbol{v}}^d_{GNSS} - \tilde{\boldsymbol{v}}^{d} = (\tilde{\boldsymbol{v}}^d_{GNSS} \times)\boldsymbol{\phi} - \boldsymbol{\delta k} \tilde{\boldsymbol{v}}^d_{GNSS}
\end{equation}
Thus, the measurement $\boldsymbol{z}$ for both M1 and M2 is:
\begin{equation}\label{eq:z_for_m1and_m2}
    \boldsymbol{z} = \tilde{\boldsymbol{v}}^d_{GNSS} - \tilde{\boldsymbol{v}}^{d}
\end{equation}
where $\tilde{\boldsymbol{v}}^{d}$ is the measured velocity from the DVL. Since the state vectors of M1 and M2 are slightly different, their corresponding measurement matrices are also varied such that:
\begin{equation}\label{eq:model_1_h_short}
    \mathbf{H}_{M1} = \begin{bmatrix}
             \tilde{\boldsymbol{v}}_{GNSS}^{d} \times & -\tilde{\boldsymbol{v}}_{GNSS}^{d}
        \end{bmatrix} \in \mathbb{R}^{3\times 4}
\end{equation}
\begin{equation}\label{eq:model_2_h_short}
    \mathbf{H}_{M2} = \begin{bmatrix}
             \tilde{\boldsymbol{v}}_{GNSS}^{d} \times & -\tilde{\boldsymbol{v}}_{GNSS,x}^{d}
        \end{bmatrix} \in \mathbb{R}^{3\times 4}
\end{equation}
where in \eqref{eq:model_2_h_short} the term $\tilde{\boldsymbol{v}}_{GNSS,x}^{d}$ is the X component of the GNSS velocity expressed in the DVL frame.
\section{Proposed Approach}\label{sec:prop_app_general}
\begin{figure}
    \centering
    \includegraphics[width=0.85\columnwidth]{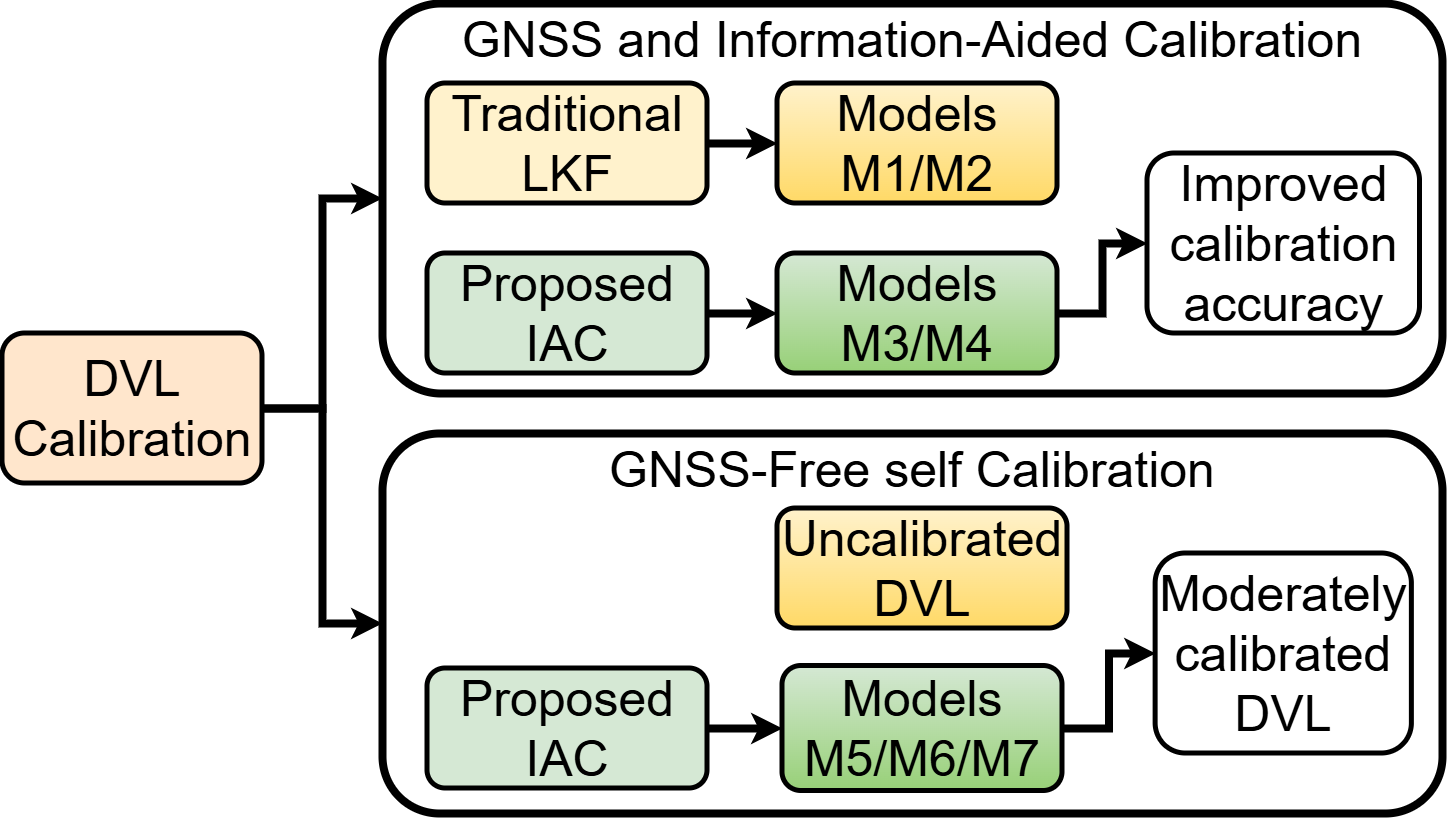}
    \caption{General diagram of our proposed IAC models (in green) in both GNSS availability scenarios.}
    \label{fig:general_diagram}
\end{figure}

In this section, we describe our five IAC models. Two IAC models (labeled M3 and M4) were designed to work alongside GNSS updates. In situations of GNSS-free conditions, that is in GNSS-denied environments or when operating without a GNSS receiver, we offer three IAC models (labeled M5, M6, and M7) for the calibration task. Figure \ref{fig:general_diagram} illustrates the proposed models, M3 to M7.
\subsection{GNSS and Information-Aided Calibration}\label{sec:gnss_enabled_IAC}
\begin{figure}
    \centering
    \includegraphics[width=0.8\columnwidth]{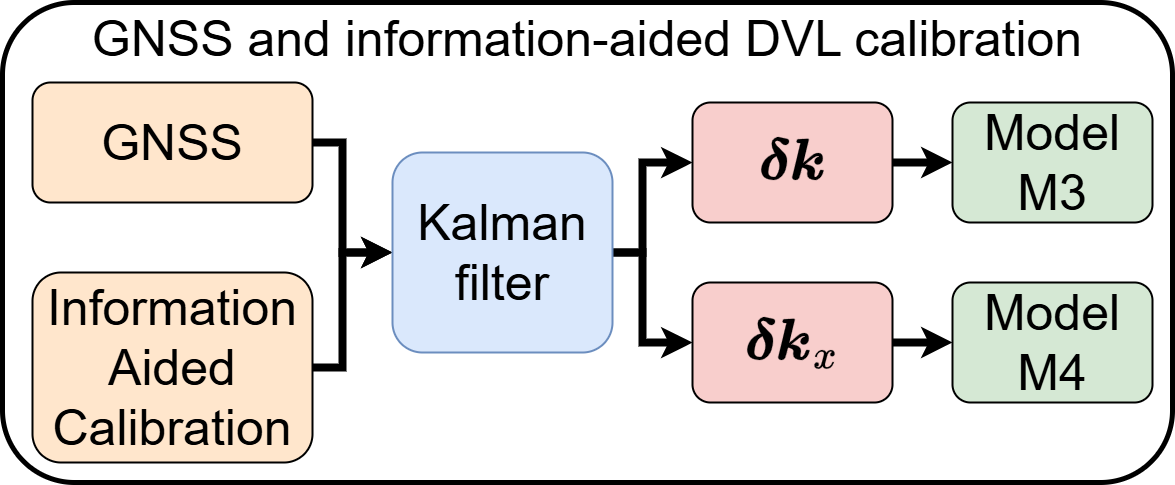}
    \caption{Overview of the proposed IAC approaches in GNSS-enabled environments showing the difference between M3 and M4.}
    \label{fig:gnss_aided_diag}
\end{figure}

Our proposed IAC approach is inspired by non-holonomic constraints commonly used in ground vehicles \cite{groves2008principles,engelsman2023information}, where no-slip and straight-line motion assumptions allow the body frame Y and Z velocity components to be approximated as zero, thereby improving accuracy and reducing drift. In this work, we adapt the non-holonomic constraints to DVL calibration and propose to apply zero velocity information aiding as an IAC in GNSS-enabled environment (IAC-G). Specifically, our proposed IAC assumes zero velocity only in the Z component of the GNSS velocity. The reason for that is twofold. First, some DVL manufacturers recommend installing the DVL with a yaw offset, causing the DVL X and Y axes to differ from those of the body frame, while the Z axis typically remains aligned. Second, the employed straight-line calibration trajectories result in nearly constant horizontal motion, Therefore, despite periodic vertical motion, the DVL Z axis velocity averages approximately to zero. Thus, our IAC-G velocity update is:
\begin{equation}\label{eq:IAC_gnss_enabled_vel}
    \tilde{\boldsymbol{v}}^{d}_{GNSS_{IAC-G}} = \begin{bmatrix}
        \tilde{\boldsymbol{v}}^{d}_{GNSS,x} \\
        \tilde{\boldsymbol{v}}^{d}_{GNSS,y} \\
        0
    \end{bmatrix}
\end{equation}
where $\tilde{\boldsymbol{v}}^{d}_{GNSS,x}$ and $\tilde{\boldsymbol{v}}^{d}_{GNSS,y}$ are the X and Y components, respectively, of the measured GNSS velocity expressed in the DVL frame. As such, we apply the proposed IAC-G non-holonomic constraints to both baseline models, M1 and M2, resulting in the following two novel models, each corresponding to one of the baseline models.
\begin{enumerate}[wide=0pt, labelwidth=!, labelindent=0pt]
\item \textbf{GNSS and IAC-G (M3)}: This model is adapted from the baseline model M1 by applying the proposed IAC-G framework. Therefore, the same state vector is utilized:
\begin{equation}\label{eq:x_m3}
    \boldsymbol{x}_{M3} = \begin{bmatrix}
        \phi_x & \phi_y & \phi_z & \delta k
    \end{bmatrix}^T
\end{equation}
M3 utilizes the same error model as in \eqref{eq:final_dvl_error_model}. However, the GNSS-measured velocity is substituted with the proposed IAC-G velocity, $\tilde{\boldsymbol{v}}^{d}_{GNSS_{IAC-G}}$, in \eqref{eq:IAC_gnss_enabled_vel}. Thus, the following measurement is obtained:
\begin{equation}\label{eq:z_m3_eq}
    \boldsymbol{z}_{M3} = \tilde{\boldsymbol{v}}_{GNSS_{IAC-G}} - \tilde{\boldsymbol{v}}^d
\end{equation}
The resulting measurement matrix is:
\begin{align}\label{eq:model_m31_h}
    &\mathbf{H}_{M3} = \begin{bmatrix}
             \tilde{\boldsymbol{v}}_{GNSS_{IAC-G}}^{d} \times & -\tilde{\boldsymbol{v}}_{GNSS_{IAC-G}}^{d}
        \end{bmatrix} = \\ \notag
        & \begin{bmatrix}
    0 & 0 & \tilde{\boldsymbol{v}}_{GNSS,y}^{d} & -\tilde{\boldsymbol{v}}_{GNSS,x}^{d} \\
   0 & 0 & -\tilde{\boldsymbol{v}}_{GNSS,x}^{d} & -\tilde{\boldsymbol{v}}_{GNSS,y}^{d}\\
    -\tilde{\boldsymbol{v}}_{GNSS,y}^{d} & \tilde{\boldsymbol{v}}_{GNSS,x}^{d} & 0 & 0
        \end{bmatrix}
\end{align}
\item \textbf{GNSS and IAC-G (M4)}: This model applies the proposed IAC-G framework to the baseline model M2 and therefore utilizes the same state vector:
\begin{equation}
    \boldsymbol{x} = \begin{bmatrix}
        \phi_x & \phi_y & \phi_z & \delta k_x
    \end{bmatrix}^T
\end{equation}
By substituting the proposed IAC-G velocity, $\tilde{\boldsymbol{v}}_{GNSS_{IAC-G}}$, from \eqref{eq:IAC_gnss_enabled_vel}, into the error model in \eqref{eq:final_dvl_error_model}, the following measurement is obtained:
\begin{equation}\label{eq:z_m4_eq}
    \boldsymbol{z}_{M4} = \tilde{\boldsymbol{v}}_{GNSS_{IAC-G}} - \tilde{\boldsymbol{v}}^d
\end{equation}
Subsequently, the measurement matrix is:
\begin{align}\label{eq:H_model_M4}
    & \mathbf{H}_{M4} = \begin{bmatrix}
             \tilde{\boldsymbol{v}}_{GNSS_{IAC-G}}^{d} \times & -\tilde{\boldsymbol{v}}_{GNSS,x}^{d}
        \end{bmatrix} =  \\ \notag
        & \begin{bmatrix}
    0 & 0 & \tilde{\boldsymbol{v}}_{GNSS,y}^{d} & -\tilde{\boldsymbol{v}}_{GNSS,x}^{d} \\
   0 & 0 & -\tilde{\boldsymbol{v}}_{GNSS,x}^{d} & 0\\
    -\tilde{\boldsymbol{v}}_{GNSS,y}^{d} & \tilde{\boldsymbol{v}}_{GNSS,x}^{d} & 0 & 0
        \end{bmatrix}
\end{align}
\end{enumerate}
Figure \ref{fig:gnss_aided_diag} illustrates the information flow of the two proposed IAC-G approaches.
\subsection{GNSS-Free Information-Aided Calibration}\label{sec:gnss_denied_calib}
\begin{figure}
    \centering
    \includegraphics[width=0.95\columnwidth]{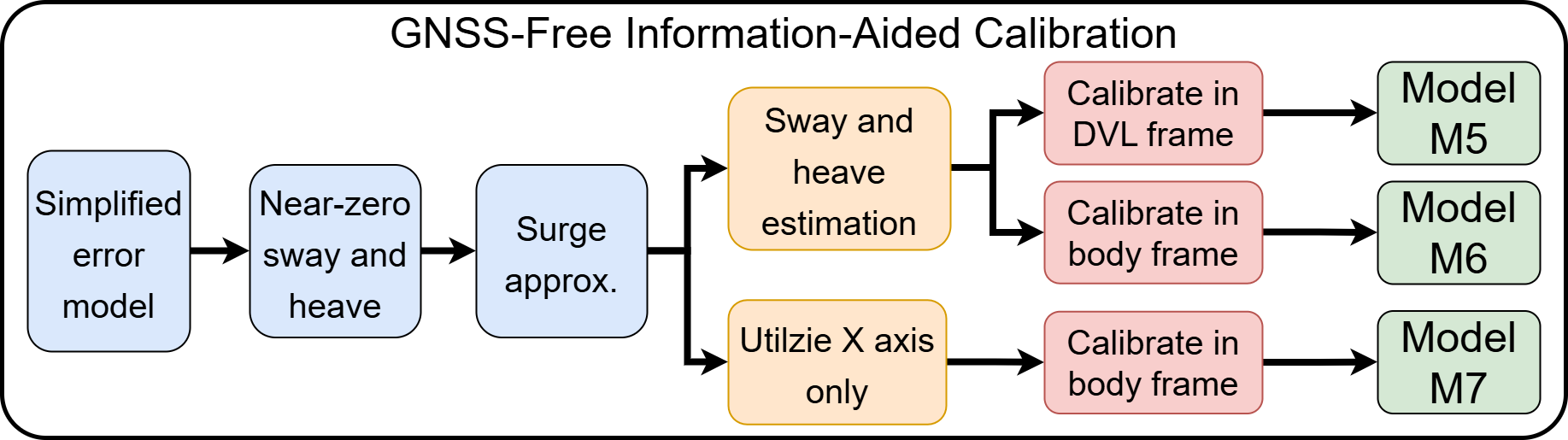}
    \caption{A block diagram illustrating the information flow for the proposed GNSS-free IAC models M5, M6, and M7.}
    \label{fig:gnss_denied_calib}
\end{figure}

\noindent This section presents the proposed GNSS-free IAC (IAC-GF) framework for GNSS-denied environments, based on the following three assumptions for GNSS-free DVL self-calibration:
\begin{enumerate}
    \item \textbf{Simplified error model}: A simple relaxed error model that employs only misalignment errors as follows:
\begin{equation}\label{eq:simplified_misalign_err_model}
            \tilde{\boldsymbol{v}}^d = (\mathbf{I} + \boldsymbol{\phi})\hat{\mathbf{T}}_{b}^{d}\tilde{\boldsymbol{v}}^b_{IAC-GF}
        \end{equation}
    where $\tilde{\boldsymbol{v}}^b_{IAC-GF}$ denotes the IAC-GF velocity vector.
    \item \textbf{Near-zero sway and heave velocity components}:
    We propose approximating the Y (sway) and Z (heave) components of the missing GNSS velocity components expressed in the body frame as small, near-zero values, such that:
        \begin{equation}\label{eq:gf_iac_x_larger_than_y_z}
        \tilde{\boldsymbol{v}}^{b}_{IAC-GF,x} >> \tilde{\boldsymbol{v}}^{b'}_{IAC-GF,y}, \tilde{\boldsymbol{v}}^{b'}_{IAC-GF,z}
    \end{equation}
   where $\tilde{\boldsymbol{v}}^{b}_{IAC-GF,x}$, $\tilde{\boldsymbol{v}}^{b}_{IAC-GF,y}$, and $\tilde{\boldsymbol{v}}^{b}_{IAC-GF,z}$ denote the unknown surge, sway, and heave components, respectively. Note that approximating or estimating the sway and heave components requires prior knowledge of the system dynamics, assumptions regarding the vehicle motion, or suitable heuristics for their estimation.   
    \item \textbf{Surge velocity approximation}:
    We propose deriving an approximation for the unknown surge velocity component, $\tilde{\boldsymbol{v}}_{IAC-GF,x}^b$, by leveraging the second assumption together with the following norm-based derivation. Since the norm operation is invariant to coordinate frame transformations, we have:
        \begin{equation}\label{eq:norm_approx_DVL_to_GNSS}
        || \tilde{\boldsymbol{v}}^d || = ||\tilde{\boldsymbol{v}}^{b}_{IAC-GF} ||
    \end{equation}
    By incorporating the second assumption in \eqref{eq:gf_iac_x_larger_than_y_z} and explicitly applying the norm operation in \eqref{eq:norm_approx_DVL_to_GNSS}, the following expression is obtained:
    \begin{equation}\label{eq:gnss_x_approx_by_dvl_norm}
        || \tilde{\boldsymbol{v}}^{d} || = || \tilde{\boldsymbol{v}}^{b}_{IAC-GF} || \approx \tilde{\boldsymbol{v}}^b_{IAC-GF,x}
    \end{equation}
    Thus, an approximation for the unknown surge component of the IAC-GF velocity expressed in the body frame is derived. Accordingly, the complete approximation of the IAC-GF velocity vector is given by:
    \begin{equation}\label{eq:iac_gf_approx_ref_vel}
        \tilde{\boldsymbol{v}}_{IAC-GF}^b = \begin{bmatrix}
            || \tilde{\boldsymbol{v}}^d || \\ \tilde{\boldsymbol{v}}^{b}_{IAC-GF,y} \\ \tilde{\boldsymbol{v}}^{b}_{IAC-GF,z}
        \end{bmatrix}
    \end{equation}
\end{enumerate} 
Using these assumptions, we propose three GNSS-free DVL self-calibration approaches, as follows.
\begin{enumerate}[wide=0pt, labelwidth=!, labelindent=0pt]
    \item \textbf{GNSS-Free IAC (M5)}: This model utilizes the simplified error model presented in \eqref{eq:simplified_misalign_err_model} with the proposed IAC-GF velocity in \eqref{eq:iac_gf_approx_ref_vel} to achieve:
\begin{equation}\label{eq:m5_error_model}
            \tilde{\boldsymbol{v}}^d = (\mathbf{I} + \boldsymbol{\phi}\times)\hat{\mathbf{T}}_{b}^{d}\tilde{\boldsymbol{v}}_{IAC-GF}^b
\end{equation}
Subsequently, the state vector is:
\begin{equation}
    \boldsymbol{x}_{M5} = \begin{bmatrix}
        \phi_x & \phi_y & \phi_z
    \end{bmatrix}^T
\end{equation}
By further deriving \eqref{eq:m5_error_model}, the following measurement is obtained:
\begin{equation}\label{eq:z_model_M5}
    \boldsymbol{z}_{M5} = \hat{\mathbf{T}}_b^d \tilde{\boldsymbol{v}}_{IAC-GF}^b - \tilde{\boldsymbol{v}}^d
\end{equation}
Subsequently, the measurement matrix is:
\begin{equation}\label{eq:H_model_M5}
    \mathbf{H}_{M5} = \begin{bmatrix}
        (\hat{\mathbf{T}}_{b}^{d}\tilde{\boldsymbol{v}}_{IAC-GF}^b) \times
    \end{bmatrix} \in \mathbb{R}^{3\times 3}
\end{equation}
Note that both the measurement $\boldsymbol{z}_{M5}$ and the measurement matrix $\mathbf{H}_{M5}$ only utilize the DVL norm and prior knowledge of the sway and heave components \eqref{eq:iac_gf_approx_ref_vel}.
\item \textbf{GNSS-Free IAC (M6)}: Here, we utilize the simplified error model in \eqref{eq:simplified_misalign_err_model} and the approximated IAC-GF velocity from \eqref{eq:iac_gf_approx_ref_vel}, but perform the calibration in the body frame instead of the DVL frame. Thus, the following is obtained:
\begin{equation}\label{eq:model_m6_m7_error_model}
    \tilde{\boldsymbol{v}}_{IAC-GF}^b = \hat{\mathbf{T}}_{d}^{b}(\mathbf{I} - \boldsymbol{\phi} \times)\tilde{\boldsymbol{v}}^{d}
\end{equation}
The state vector is defined as:
\begin{equation}\label{eq:x_of_M6}
    \boldsymbol{x}_{M6} = \begin{bmatrix}
        \phi_x & \phi_y & \phi_z
    \end{bmatrix}^T
\end{equation}
The measurement is derived as follows:
\begin{equation}\label{eq:m6_measurement}
    \boldsymbol{z}_{M6} =  \tilde{\boldsymbol{v}}_{IAC-GF}^b - \hat{\mathbf{T}}_{d}^{b} \tilde{\boldsymbol{v}}^{d}
\end{equation}
Subsequently, the measurement matrix is obtained:
\begin{equation}
    \mathbf{H}_{M6} = \begin{bmatrix}
        \hat{\mathbf{T}}_{d}^{b} (\tilde{\boldsymbol{v}}^d \times)
    \end{bmatrix}
\end{equation}
Note that, since the calibration is performed in the body frame, the measurement matrix is a function of the measured DVL velocity, unlike in M5, where it is a function of the approximated IAC-GF velocity.
\item \textbf{GNSS-Free IAC (M7)}: 
By employing the same error model as M6 in \eqref{eq:model_m6_m7_error_model}, the state vector is defined as:
\begin{equation}
    \boldsymbol{x}_{M7} = \begin{bmatrix}
        \phi_x & \phi_y & \phi_z
    \end{bmatrix}^T
\end{equation}
To achieve the proposed IAC-GF DVL self-calibration without any prior knowledge of the sway and heave components required by M5 and M6, we modify the measurement in \eqref{eq:m6_measurement} and utilize only its first component, as follows:
\begin{align}\label{eq:z_M7}
    \boldsymbol{z}_{M7} & = \begin{bmatrix}
        1 & 0 & 0
    \end{bmatrix} (\tilde{\boldsymbol{v}}_{IAC-GF}^b - \hat{\mathbf{T}}_{d}^{b}\tilde{\boldsymbol{v}}^{d})^T \\ \notag
    & = ||\tilde{\boldsymbol{v}}_{d} || - (\hat{\mathbf{T}}_{d}^{b}\tilde{\boldsymbol{v}}^{d})_{x}
\end{align}
where $(\hat{\mathbf{T}}_{d}^{b}\tilde{\boldsymbol{v}}^{d})_{x}$ is the X axis component of the measured DVL velocity expressed in the body frame. Similarly, the measurement matrix is defined as:
\begin{equation}\label{eq:H_m7}
    \mathbf{H}_{M7} = \begin{bmatrix}
        1 & 0 & 0
    \end{bmatrix} \begin{bmatrix}
        \hat{\mathbf{T}}_{d}^{b} (\tilde{\boldsymbol{v}}^d \times)
    \end{bmatrix}  \in \mathbb{R}^{1 \times 3}
\end{equation}
Thus, M7 utilizes only DVL measurements, without requiring prior knowledge or approximations of the sway and heave components, thereby enabling a novel GNSS-free DVL self-calibration paradigm.
\end{enumerate}
Figure \ref{fig:gnss_denied_calib} illustrates the three proposed IAC approaches for GNSS-free DVL self-calibration.
\subsection{IAC Approaches Summary}\label{sec:iac_summary}
The proposed IAC-G and IAC-GF approaches result in five different linear Kalman filter models for GNSS-enabled and GNSS-free environments, respectively. In GNSS-enabled environments, models M3 and M4 employ the proposed IAC-G framework together with a four-state vector consisting of misalignment errors and a scale factor. In GNSS-free environments, models M5, M6, and M7 employ a simplified error model together with the proposed IAC-GF framework, subsequently utilizing a three-state vector consisting only of misalignment errors. Models M5 and M6 require prior information to enable GNSS-free calibration, whereas M7 provides a fully standalone GNSS-free DVL self-calibration approach. Each of the five proposed models employs a different IAC assumption based on either the GNSS reference velocity, when available, or the approximated IAC-GF velocity as the reference. Accordingly, the models differ in their measurement vectors and measurement matrices. Table \ref{tbl:IAC_models_prop} summarizes the properties of the five proposed IAC models.
\begin{table*}[h!]
\centering
\caption{Proposed IAC model properties including aiding type, modified GNSS velocity measurement, state vector, measurement, and measurement matrix.}\label{tbl:IAC_models_prop}
\resizebox{0.99\textwidth}{!}{%
\begin{tabular}{|c|c|c|c|c|c|c|}
\hline
 Model  & \begin{tabular}[c]{@{}c@{}}IAC\\ Type\end{tabular}       & \begin{tabular}[c]{@{}c@{}}Baseline\\ Model\end{tabular} & \begin{tabular}[c]{@{}c@{}}IAC\\ Measurement\end{tabular} & \begin{tabular}[c]{@{}c@{}}State\\ Vector\end{tabular} & Measurement & \begin{tabular}[c]{@{}c@{}}Measurement\\ Matrix\end{tabular} \\ \hline
M3 & \begin{tabular}[c]{@{}c@{}}GNSS-enabled \end{tabular} & M1                                                       & $\tilde{\boldsymbol{v}}_{GNSS_{IAC-G}}^d$                                                                  & $\begin{bmatrix} \phi_x & \phi_y & \phi_z & \delta k \end{bmatrix}^T$                                                   &     $\tilde{\boldsymbol{v}}_{GNSS_{IAC-G}}^d$ - $\tilde{\boldsymbol{v}}^d$     & $\begin{bmatrix} \tilde{\boldsymbol{v}}_{GNSS_{IAC-G}}^d \times & -\tilde{\boldsymbol{v}}_{GNSS_{IAC-G}}^d \end{bmatrix} \in \mathbb{R}^{3\times4}$                                                          \\ \hline
M4 & \begin{tabular}[c]{@{}c@{}}GNSS-enabled \end{tabular} & M2                                                       & $\tilde{\boldsymbol{v}}_{GNSS_{IAC-G}}^d$                                                                    & $\begin{bmatrix} \phi_x & \phi_y & \phi_z & \delta k_x \end{bmatrix}^T$                                                    & $\tilde{\boldsymbol{v}}_{GNSS_{IAC-G}}^d$ - $\tilde{\boldsymbol{v}}^d$         & $\begin{bmatrix} \tilde{\boldsymbol{v}}_{GNSS_{IAC-G}}^d \times & -\tilde{\boldsymbol{v}}_{GNSS_{IAC-G},x}^d \end{bmatrix}\in \mathbb{R}^{3\times4}$                                                          \\ \hline
M5 & \begin{tabular}[c]{@{}c@{}}GNSS-free\end{tabular}     & Uncalibrated                                             & $\tilde{\boldsymbol{v}}^b_{IAC-GF}$  &         $\begin{bmatrix} \phi_x & \phi_y & \phi_z\end{bmatrix}^T$                                            & $\hat{\mathbf{T}}_b^d \tilde{\boldsymbol{v}}^b_{IAC-GF}$ - $\tilde{\boldsymbol{v}}^d$       & $\begin{bmatrix} (\hat{\mathbf{T}}_{b}^{d}\tilde{\boldsymbol{v}}_{IAC-GF}^b) \times \end{bmatrix}\in \mathbb{R}^{3\times3}$                                                          \\ \hline
M6 & \begin{tabular}[c]{@{}c@{}}GNSS-free\end{tabular}     & Uncalibrated                                             & $\tilde{\boldsymbol{v}}^b_{IAC-GF}$                                                                  &  $\begin{bmatrix} \phi_x & \phi_y & \phi_z\end{bmatrix}^T$                                                    & $\tilde{\boldsymbol{v}}^b_{IAC-GF}$  - $\hat{\mathbf{T}}_d^b \tilde{\boldsymbol{v}}^d$        & $\begin{bmatrix} \hat{\mathbf{T}}_d^b (\tilde{\boldsymbol{v}}^d \times ) \end{bmatrix}\in \mathbb{R}^{3\times3}$                                                           \\ \hline
M7 & \begin{tabular}[c]{@{}c@{}}GNSS-free\end{tabular}     & Uncalibrated                                             & $\tilde{\boldsymbol{v}}^b_{IAC-GF}$                                                                  &  $\begin{bmatrix} \phi_x & \phi_y & \phi_z\end{bmatrix}^T$                                                    & $|| \tilde{\boldsymbol{v}}_d|| - (\hat{\mathbf{T}}_d^b \tilde{\boldsymbol{v}}^d)_{x}$         & $\begin{bmatrix}
    1 & 0 & 0 
\end{bmatrix}$ $\begin{bmatrix}
    \hat{\mathbf{T}}_d^b (\tilde{\boldsymbol{v}}^d \times)
\end{bmatrix}\in \mathbb{R}^{1\times3}$                                                          \\ \hline
\end{tabular}%
}
\end{table*}
\section{Analysis and Results}\label{sec:res_gen}
\subsection{Dataset}\label{sec:dataset}
\begin{figure}
\centering

\subfloat[T12]{%
    \includegraphics[width=0.9\linewidth]{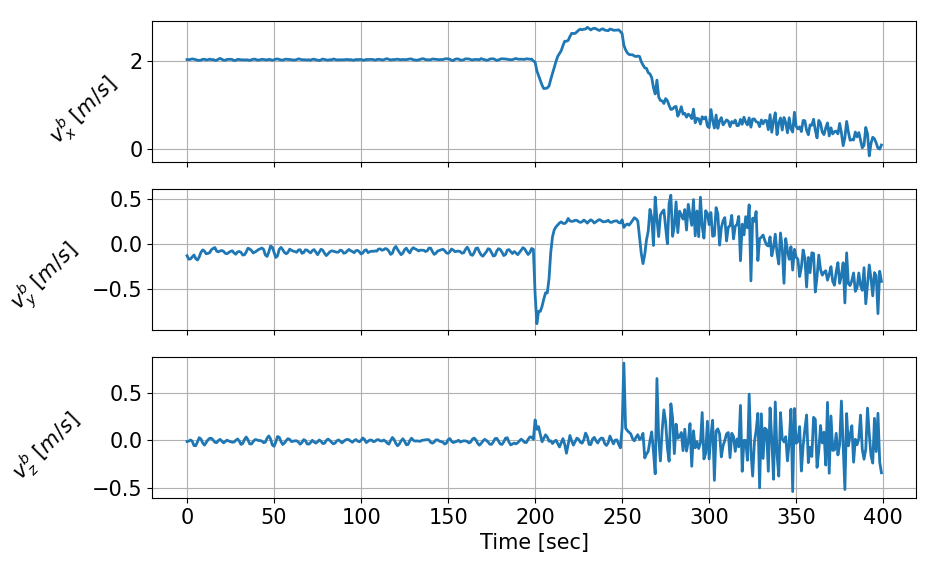}
    \label{fig:t11_plot}
}

\vspace{0.5em}

\subfloat[T14]{%
    \includegraphics[width=0.9\linewidth]{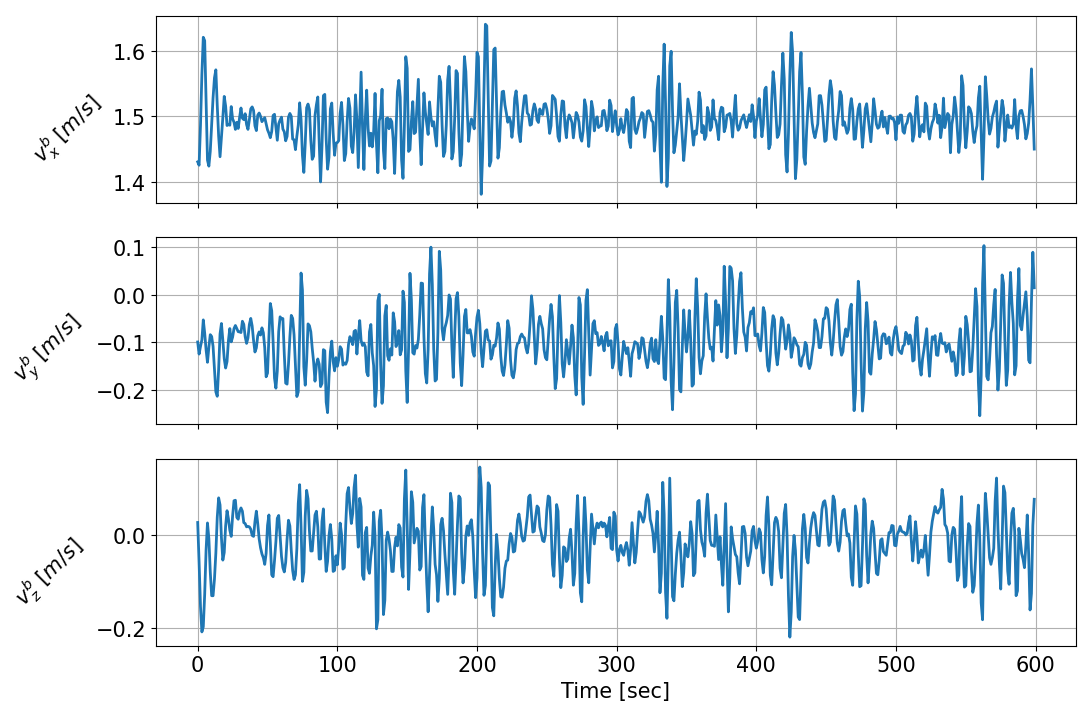}
    \label{fig:t14_plot}
}

\caption{DVL velocity vector expressed in the body frame, from Trajectory T12 (first dataset) and Trajectory T14 (second dataset).}
\label{fig:t11_t14_vel_plots}
\end{figure}

To validate the proposed IAC approaches, two real-world datasets collected in the Mediterranean Sea using the University of Haifa AUV Snapir \cite{snapir_link} were utilized. The first dataset contains thirteen trajectories, T1 to T13, each with a duration of 400 seconds and with different AUV dynamics \cite{cohen2025adaptive}. The total duration of the first dataset is approximately 86 minutes. The second dataset includes two additional trajectories, T14 and T15, with durations of 500 and 600 seconds, respectively \cite{cohen2022beamsnet}. Thus, a total of fifteen recorded trajectories are used for the evaluation. To illustrate the recorded velocity dynamics, Figure \ref{fig:t11_t14_vel_plots} shows the recorded DVL velocity in both frames for trajectories T11 and T14.
\subsection{Validation Procedure}\label{sec:valid_procedure}
\begin{figure}
    \centering
    \includegraphics[width=0.95\linewidth]{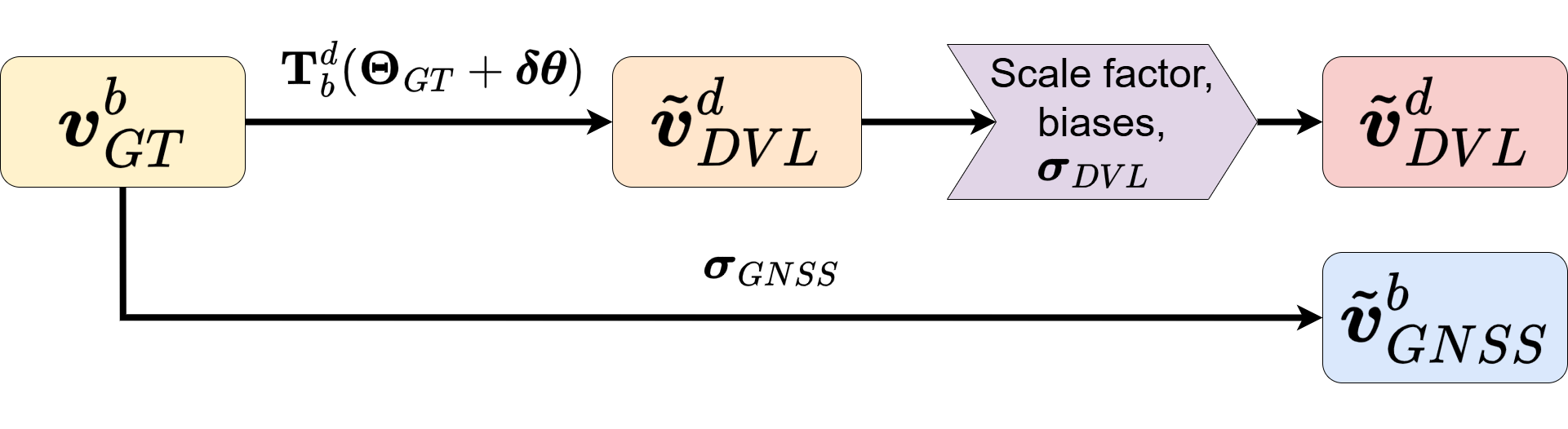}
    \caption{Block diagram of the noising pipeline applied to the GT DVL measurements to generate noised DVL measurements with misalignment errors and the reference GNSS measurements.}
    \label{fig:noising_pipeline}
\end{figure}

To sufficiently evaluate the accuracy of the proposed IAC approaches against the baseline methods, a noising pipeline was employed for two purposes. First, common DVL error terms were introduced to the recorded DVL measurements to generate unit-under-test (UUT) noised DVL measurements representative of a real-world calibration scenario. Second, since neither dataset includes GNSS velocity measurements, reference GNSS velocity measurements were generated for the evaluation process.\\
To generate the UUT DVL measurements, the error terms applied to the DVL include scale-factor, biases, zero-mean Gaussian white noise, and misalignment errors between the DVL and body frames. These error terms are common DVL error sources and are representative of the characteristics of real-world DVL measurements \cite{yampolsky2025dcnet,damari2026resalignnet,li2022calibration,dvl_a50}. To generate the reference GNSS velocity measurements, only zero-mean Gaussian white noise was added to the GT velocity. The noising pipeline, together with the corresponding DVL and GNSS error models, was adopted from \cite{yampolsky2025dcnet}. It is important to note that the generated UUT DVL measurements are derived from real-world DVL data collected during sea experiments rather than from simulated data. Consequently, the validation trajectories preserve the characteristics and dynamics of real-world operations. Figure \ref{fig:noising_pipeline} illustrates the employed noising pipeline.
There, the velocity $\boldsymbol{v}_{GT}^b$ corresponds to the ground truth (GT) DVL velocity vector. In addition, the misalignment errors are introduced indirectly through small Euler-angle perturbations, denoted by $\boldsymbol{\delta \theta}$. The values selected for the error terms utilized in the noising pipeline to generate the UUT DVL measurements are presented in Table \ref{tbl:error_terms_used}, which includes five Euler-angle perturbation variations, three scale-factor variations, one DVL bias variation, and one DVL and GNSS zero-mean Gaussian white noise standard deviation (STD) value. 
\begin{table}[h!]
\centering
\caption{Error term combinations used in the validation procedure.}\label{tbl:error_terms_used}
\resizebox{0.95\columnwidth}{!}{%
\begin{tabular}{|c|c|c|}
\hline
  \begin{tabular}{c}   Error term \\ parameter   \end{tabular}      &\begin{tabular}{c} Error term \\ combination values \end{tabular} & \begin{tabular}[c]{c} Number of \\ combinations
             \end{tabular} \\ \hline
$\boldsymbol{\delta \theta}$ $[\degree]$  & \begin{tabular}{c} $[5,-0.5,2.5]$, $[0.5,1,-1.5]$,\\ $[-2.5,1.8,4]$, $[-1,-2.5,-3]$,\\$[2,0.5,4.2]$\end{tabular} & 5                     \\ \hline
$\boldsymbol{\delta k}$ $[\%]$ & \begin{tabular}{c} $[0.5,0.3,0.7]$, $[0.2,0.7,0.6]$,\\$[0.9,0.5,0.1]$\end{tabular}                & 3                     \\ \hline
$\boldsymbol{b}_{DVL}$ $[m/s]$        & $[0.007,0.001,0.003,0.009]$                                                              & 1                     \\ \hline
$\boldsymbol{\sigma}_{DVL}$ $[m/s]$ & $5\cdot 10^{-3}$                                                                                          & 1                     \\ \hline
$\boldsymbol{\sigma}_{GNSS}$ $[m/s]$  & $0.2$                                                                                        & 1                     \\ \hline
\end{tabular}%
}
\end{table}\\
By combining the five Euler-angle perturbation variations with the three scale-factor variations, fifteen different DVL error-term combinations are obtained and evaluated. Although the DVL bias variation and DVL noise STD value do not contribute to the number of combinations, they are included in all evaluations to ensure that the generated UUT DVL measurements reflect realistic DVL measurement characteristics. For the GNSS measurements, only a single zero-mean Gaussian white noise STD value is considered, since the primary focus of this study is the evaluation of DVL calibration under different DVL error conditions, particularly scale-factor errors and DVL-to-body-frame misalignment.\\
To thoroughly evaluate the proposed IAC approaches, a cross-validation-based evaluation procedure based on four evaluation sets was employed. The complete dataset of fifteen trajectories was divided into four evaluation sets, E1 to E4, each defined by a different calibration trajectory. In each set, the selected calibration trajectory is used to estimate the DVL error terms, while the remaining fourteen trajectories are used for evaluation after being calibrated with the estimated error terms. The four calibration trajectories are denoted as C1 to C4 and correspond to T12, T13, T14, and T15, respectively. They were selected because they, or segments of their recorded data, exhibit straight-line, near-constant-velocity dynamics, which constitute the required calibration trajectory dynamics in this work. Consequently, for each evaluation set, the fourteen remaining trajectories consist of all trajectories except the selected calibration trajectory. Table \ref{tbl:eval_proc_info} provides a detailed description of the four evaluation sets, including their calibration and evaluation trajectories.
\begin{table}[!h]
\centering
\caption{Calibration and evaluation trajectories in each evaluation set.}\label{tbl:eval_proc_info}
\resizebox{\columnwidth}{!}{%
\begin{tabular}{|c|c|c|c|}
\hline
\begin{tabular}[c]{@{}c@{}}Evaluation\\ set\end{tabular} & 
\begin{tabular}[c]{@{}c@{}}Calib.\\ trajectory\end{tabular} & \begin{tabular}[c]{@{}c@{}}Utilized calib.\\ time window\end{tabular} & \begin{tabular}[c]{@{}c@{}}Evaluated \\ trajectories\end{tabular} \\ \hline
E1 & C1                                                         & 0-200  [sec]                                                                                                                                 & T1 to T11, C2, C3, C4                                                                            \\ \hline
E2 & C2                                                         & All                                                                                                                                     & T1 to T11, C1, C3, C4                                                                            \\ \hline
E3 & C3                                                         & All                                                                                                                                     & T1 to T11, C1, C2, C4                                                                            \\ \hline
E4 & C4                                                         & All                                                                                                                                    & T1 to T11, C1, C2, C3                                                                            \\ \hline
\end{tabular}%
}
\end{table} 
Note that when C1, corresponding to T12, is selected as the calibration trajectory, only its first 200 seconds are utilized. This is because only the first 200 seconds of T12 exhibit the straight-line dynamics required by the calibration procedure. This can be visually observed in Figure \ref{fig:t11_plot}.\\
Using the four evaluation sets and the fifteen DVL error-term combinations, the following evaluation procedure is performed:
\begin{enumerate}[topsep=0pt,itemsep=0pt,parsep=0pt,partopsep=0pt]
    \item \textbf{Step 1}: Select one of the four evaluation sets, E1 to E4.
    \item \textbf{Step 2}: Select an error-term combination (Table {\ref{tbl:error_terms_used}}) and apply it to all 15 trajectories.
    \item \textbf{Step 3}: Estimate the DVL error terms using the calibration trajectory associated with the evaluation set selected in \textbf{Step 1}.
    \item \textbf{Step 4}: Use the estimated DVL error terms to calibrate the remaining 14 trajectories.
    \item \textbf{Step 5}: Evaluate the method performance by comparing the calibrated velocities with the GT velocities.
    \item \textbf{Step 6}: Repeat Steps 2 - 5 for all error-term combinations.
    \item \textbf{Step 7}: Repeat Steps 1 - 6 for all evaluation sets.
\end{enumerate}
The above evaluation procedure is performed for all approaches, including the baseline models, M1 and M2, and the proposed IAC approaches, M3 to M7. The average velocity values of the four calibration trajectories, expressed in the body frame, are presented in Table \ref{tbl:true_vel_against_used}. The diversity in their average velocity and duration enables a comprehensive assessment of the robustness and performance of the different approaches under a wide range of operating conditions.
\begin{table}[h!]
\centering
\caption{Calibration trajectories with average values of the velocity components.}\label{tbl:true_vel_against_used}
\resizebox{\columnwidth}{!}{%
\begin{tabular}{|c|c|c|c|}
\hline
\begin{tabular}[c]{@{}c@{}}Calib.\\ Trajectory\end{tabular} & \begin{tabular}[c]{@{}c@{}}Average  Surge\\ Velocity $[m/s]$\end{tabular} & \begin{tabular}[c]{@{}c@{}}Average Sway\\ Velocity $[m/s]$\end{tabular} & \begin{tabular}[c]{@{}c@{}}Average Heave\\ Velocity $[m/s]$\end{tabular} \\ \hline
C1                                                         & 2.04                                                                    & -0.08                                                                 & -0.01                                                                 \\ \hline
C2                                                         & 2.07                                                                    & -0.17                                                                 & -0.01                                                                 \\ \hline
C3                                                         & 1.49                                                                    & -0.1                                                                  & -0.02                                                                 \\ \hline
C4                                                         & 1.49                                                                    & -0.13                                                                 & -0.01                                                                 \\ \hline
\end{tabular}%
}
\end{table}
However, for the IAC-GF approaches, an approximation of the sway and heave velocity components is still required, as described in \eqref{eq:iac_gf_approx_ref_vel}. To this end, the sway and heave components are approximated by averaging the first two seconds of the calibration trajectory, and the resulting average values are used during the calibration process in Step 3 of the evaluation procedure. This simple heuristic enables a true end-to-end GNSS-free DVL self-calibration procedure, as it eliminates the need for any external reference velocity or additional external information during calibration.
\subsection{Performance Metrics}\label{sec:vrmse_matrix_sec}
To evaluate the calibration accuracy of each approach, first the velocity root mean squared error (VRMSE) is calculated for each one of the fourteen calibrated evaluation trajectories mentioned in Step 5 of the evaluation procedure. The VRMSE is defined as:
\begin{equation}\label{eq:vrmse_gen}
    \text{VRMSE}(\boldsymbol{v}^b_{j}, \hat{\boldsymbol{v}}^{b}_{j}) = \sqrt{ \frac{1}{N} \sum_{i=1}^{N}{ || \boldsymbol{v}^b_{i,j} - \tilde{\boldsymbol{v}}^b_{i,j} || }^2}
\end{equation}
where $\boldsymbol{v}^b_{j}$ is the GT velocity of evaluation trajectory $j$, and $\hat{\boldsymbol{v}}^{b}$ is the corresponding calibrated velocity obtained using the estimated error terms. To assess the calibration accuracy of a given method over a complete evaluation cycle, that is, after completing Steps 2 to 6 and evaluating all error-term combinations for a given evaluation set, the average VRMSE (Avg.VRMSE) is calculated as follows:
\begin{align}\label{eq:avg_vrmse_epoch}
    \text{Avg.VRMSE}_{E_e} & = \\ \notag = & \frac{1}{15}\sum_{k=1}^{15}[ \frac{1}{14}\sum_{j}^{14}\text{VRMSE}(\boldsymbol{v}^b_{j}, \hat{\boldsymbol{v}}^{b}_{j})]
\end{align}
where $\text{Avg.VRMSE}_{E_e}$ is the average VRMSE obtained for evaluation set $E_e$, with $E_e \in \{E1,E2,E3,E4\}$. The index $j$ denotes one of the fourteen evaluation trajectories considered in Step 5, while $k$ denotes the evaluated error-term combination from Step 2. Therefore, \eqref{eq:avg_vrmse_epoch} represents the performance metric of a given method for a specific evaluation set.
\subsection{GNSS IAC Results}\label{sec:IAC_gnss_enabled}
\begin{figure*}
    \centering
    \begin{adjustbox}{width = 0.99\textwidth}
    \subfloat[Model M3]{%
        \includegraphics[width=0.49\textwidth]{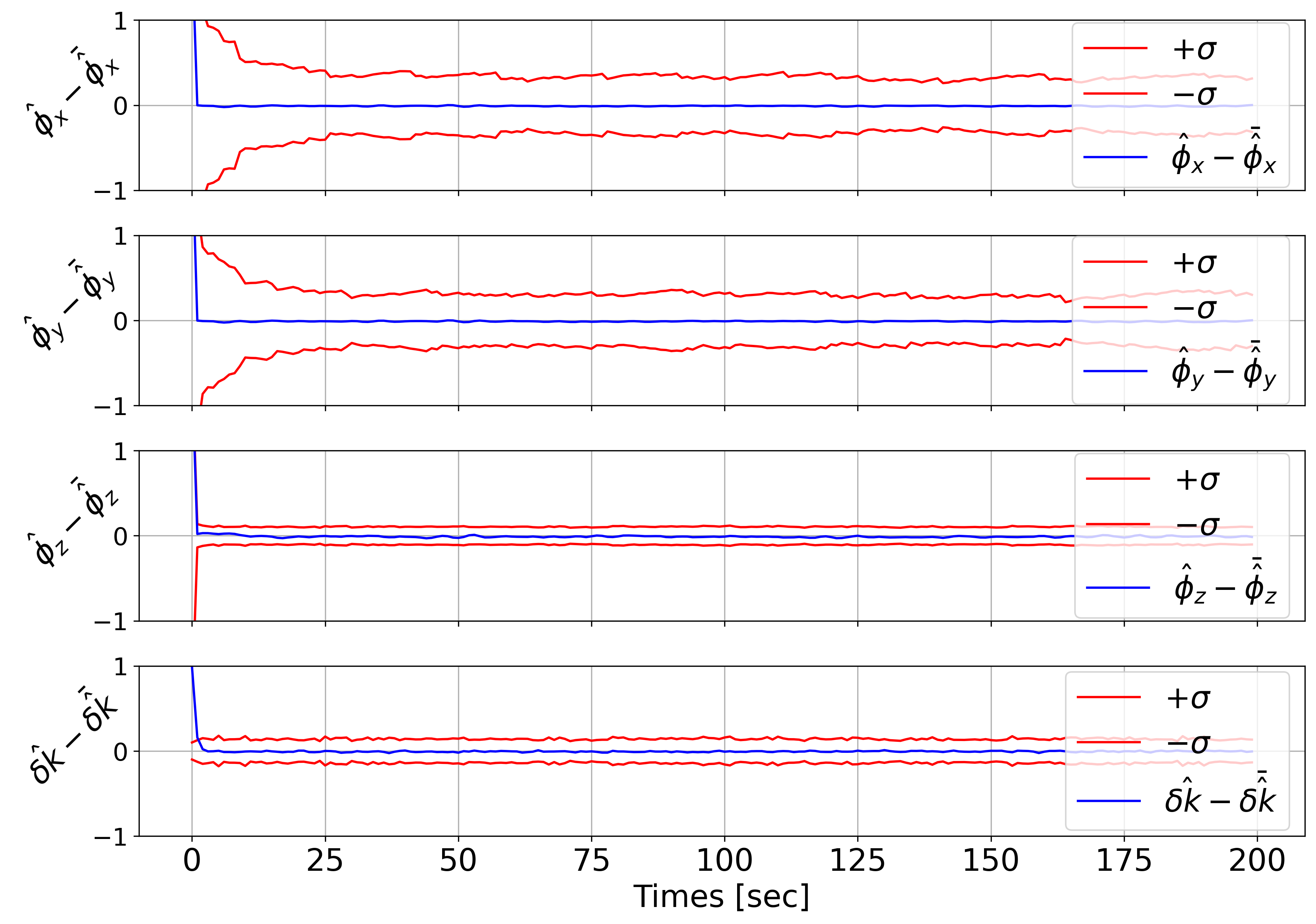}
        \label{fig:m3_cov}
    }
    \hfill
    \subfloat[Model M4]{%
        \includegraphics[width=0.49\textwidth]{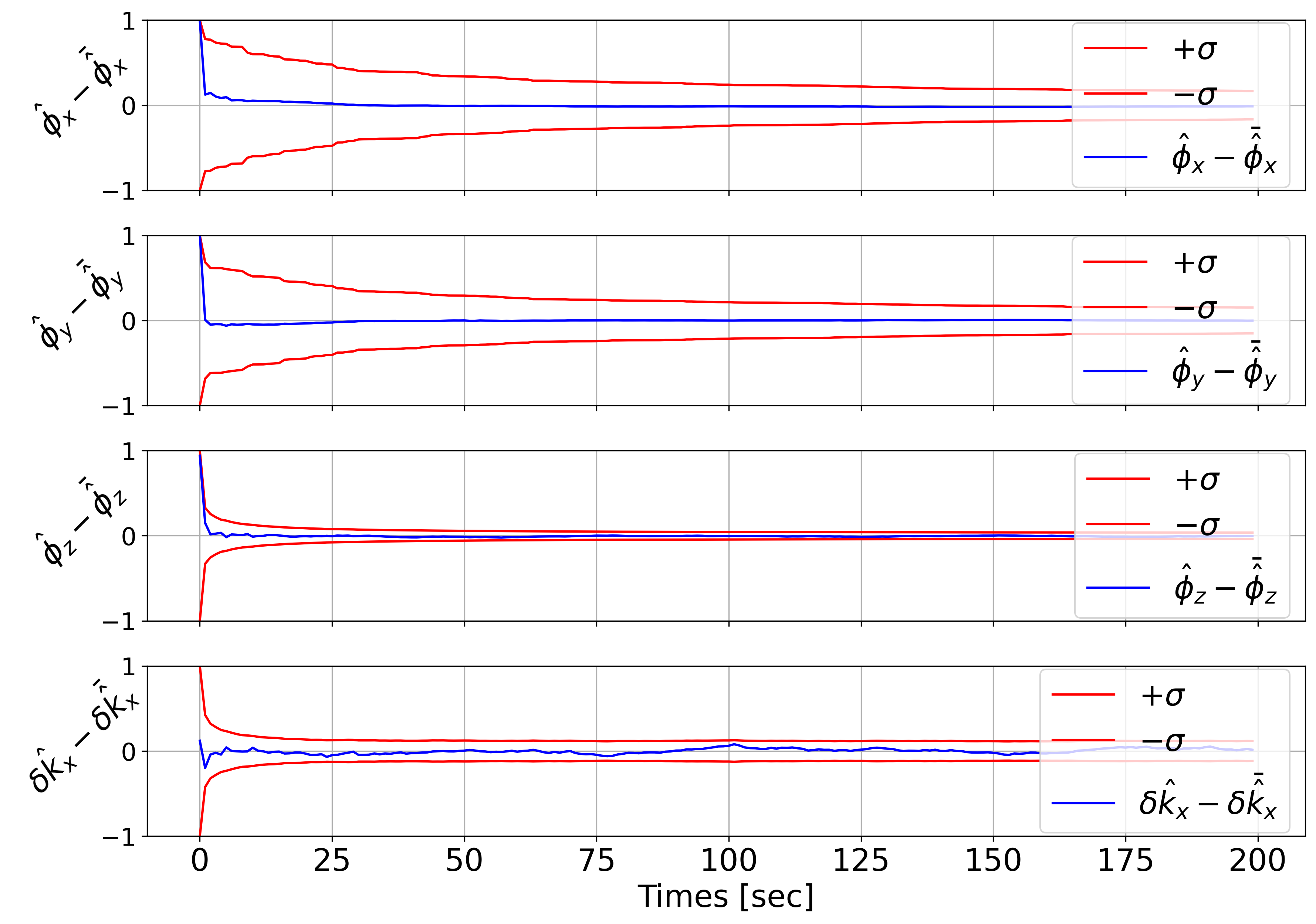}
        \label{fig:m4_cov}
    }
    \end{adjustbox}
    \caption{State estimation of models M3 (left) and M4 (right) during calibration using trajectory C1. In both figures, the blue curves represent the state estimation error, computed as the difference between the estimated state and its average value, while the red curves represent the square root of the estimated state error covariance, corresponding to the covariance sleeve. In both figures, the first three rows show the estimated misalignment errors, $\phi_x$, $\phi_y$, and $\phi_z$, while the last row shows the estimated scale factor, $\delta k$ for M3, and $\delta k_x$ for M4.}
    \label{fig:covariance_compare}
\end{figure*}

\begin{figure}
    \centering
    \includegraphics[width=0.9\columnwidth]{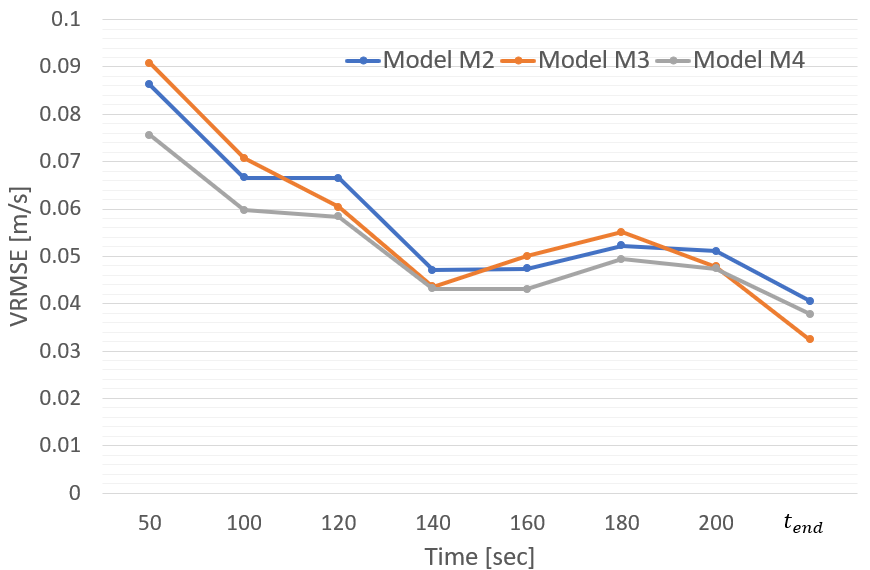}
    \caption{The average VRMSE over all calibration trajectories averaged across all fifteen error terms combinations.}
    \label{fig:IAC_gnss_aided_vrmse_as_time}
\end{figure}

First, we evaluate the proposed M3 and M4 linear Kalman filter models performance in terms of the estimated state error and the corresponding single standard deviation covariance sleeve. Figure \ref{fig:covariance_compare} presents the covariance sleeve and estimation error for each state in M3 and M4.
It shows that the state estimation error remains well within the error-state covariance sleeve when using the 200-second calibration segment of C1. Similar behavior was observed for the other calibration trajectories, C2 to C4, corresponding to evaluation sets E2 to E4, and is therefore omitted for brevity. Table \ref{tbl:IAC_gnss_eanbled_vrmse_res} presents the Avg.VRMSE values, in $[m/s]$, obtained from the validation procedure for each of the four evaluation sets using \eqref{eq:avg_vrmse_epoch}.
\begin{table}[h!]\centering
\caption{Average VRMSE values, in $[m/s]$, obtained for evaluation sets E1 to E4 using the proposed validation procedure. Numeric values in parentheses indicate the improvement relative to M2. The calibration trajectory associated with each evaluation set is indicated in parentheses as C1 to C4 in parentheses.}\label{tbl:IAC_gnss_eanbled_vrmse_res}
\resizebox{\columnwidth}{!}{%
\begin{tabular}{|c|c|c|c|c|}
\hline
  \backslashbox{Eval. set\\(Calib. traj.)}{Model}      & M1     & M2     & M3 (Ours) & M4 (Ours) \\ \hline
E1 (C1) $[m/s]$  & 0.0522 & 0.0421 & 0.0342    & 0.04    \\ \hline
E2 (C2)  $[m/s]$  & 0.0416 & 0.0396 & 0.0294    & 0.0371    \\ \hline
E3  (C3) $[m/s]$  & 0.0550 & 0.0391 & 0.0304    & 0.0328    \\ \hline
E4 (C4)  $[m/s]$  & 0.0529 & 0.0411 & 0.0564    & 0.0413    \\ \hline
Average $[m/s]$& 0.0504  & 0.0405 & \textbf{0.0323} (\textbf{20.12} $\%$)   & \textbf{0.0378} (\textbf{6.65} $\%$)    \\ \hline
\end{tabular}%
}
\end{table}
From Table \ref{tbl:IAC_gnss_eanbled_vrmse_res}, three main observations can be made. First, baseline model M2 achieves better performance than M1. Therefore, the improvements of the proposed approaches, shown in parentheses, are reported relative to M2. Second, both proposed IAC approaches, M3 and M4, provide more accurate calibration on average. Third, M3 achieves the best overall improvement, with an average accuracy gain of $20.12\%$. Fourth, both M3 and M4 achieve an average improvement of $13.4\%$ compared to the best baseline M2. Furthermore, when comparing each IAC model to its corresponding baseline, M3 improves by an average of $35.8\%$ over M1, while M4 improves by $6.7\%$ over M2, demonstrating the effectiveness of the proposed IAC. A key observation is that the only modification introduced in M3 and M4 relative to M1 and M2 was setting the Z velocity component to zero, without altering any Kalman filter parameters, further highlighting the robustness and applicability of the proposed approach.\\
To further evaluate the effectiveness of the proposed IAC models, the validation procedure was repeated while controlling the allowed calibration duration. The following calibration durations were analyzed: $t_{calib} = [50,100,120,140$ $,160,180,200,t_{end}] \: [\mathrm{sec}]$, where $t_{calib}=200 \: [\mathrm{sec}]$ corresponds to the full calibration duration of C1, but not of the other three trajectories, C2 to C4. Therefore, $t_{calib}=t_{end}$ denotes using the entire available calibration trajectory, which corresponds to the results presented in Table \ref{tbl:IAC_gnss_eanbled_vrmse_res}. Figure \ref{fig:IAC_gnss_aided_vrmse_as_time} presents the average Avg.VRMSE for each allowed calibration duration, obtained by averaging the Avg.VRMSE values corresponding to the four evaluation sets. Only baseline model M2 is shown as baseline, as Table \ref{tbl:IAC_gnss_eanbled_vrmse_res} demonstrated its superiority over M1.
From Figure \ref{fig:IAC_gnss_aided_vrmse_as_time}, three main observations can be made. First, all three models, M2, M3, and M4, demonstrate a convergence behaviour in which the VRMSE decreases as the calibration time increases. Second, M4 consistently outperforms its baseline model, M2. Third, M3 achieves the best results for the longest allowed calibration time, that is, $t_{calib}=t_{end}$, which differs between the calibration trajectories. However, when considering a fixed allowed calibration time of $200$ seconds, it is evident that M3 and M4 outperform the baseline M2 and achieve better results. Overall, the proposed IAC approaches improve calibration accuracy, with M3 providing the best performance in both accuracy and calibration time reduction.
\subsection{GNSS-Free IAC Results}\label{sec:ZVA_GD_results}
\begin{figure*}
    \centering
    \subfloat[Model M5]{%
        \includegraphics[width=0.32\textwidth]{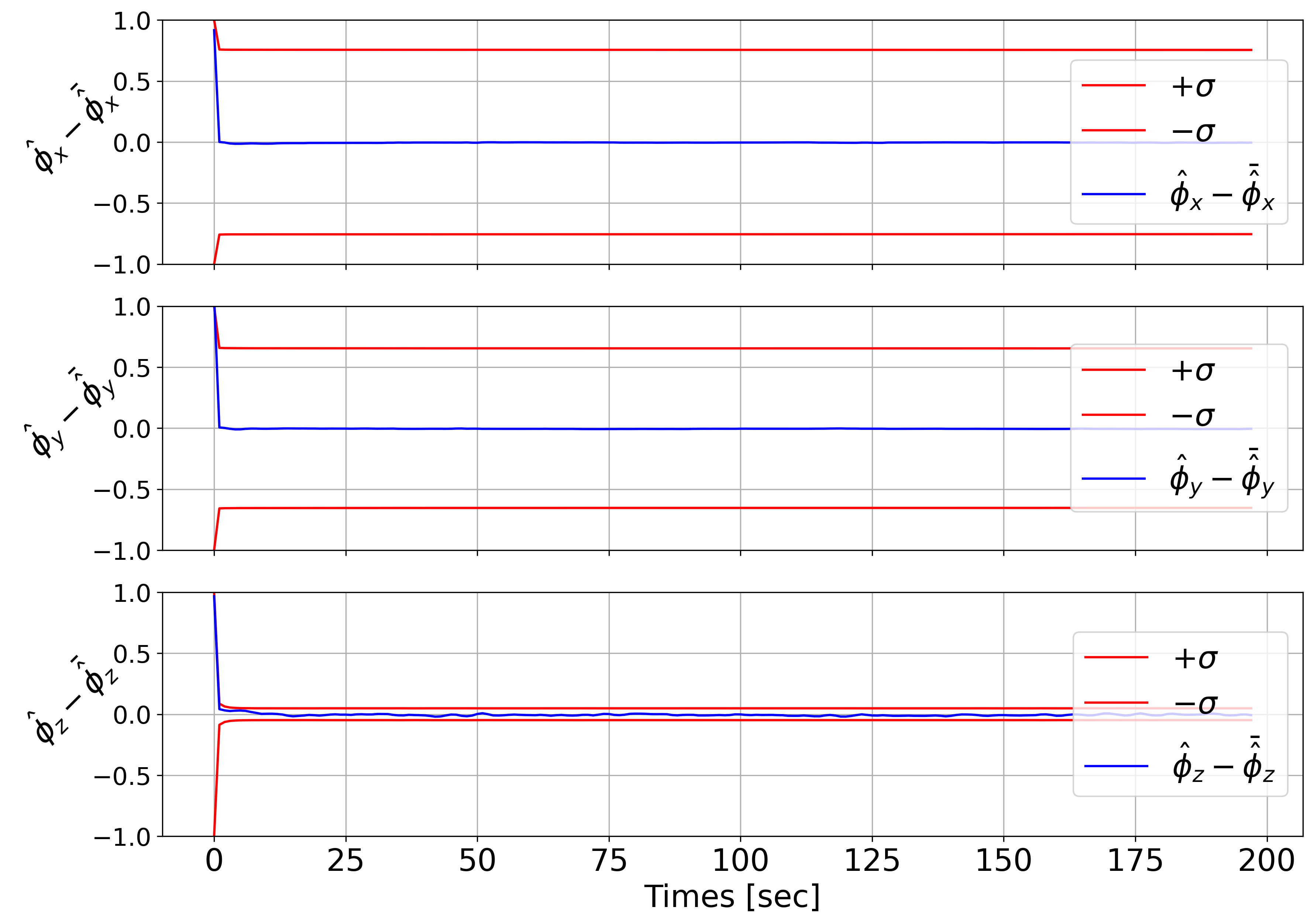}
        \label{fig:m5_cov}
    }
    \hfill
    \subfloat[Model M6]{%
        \includegraphics[width=0.32\textwidth]{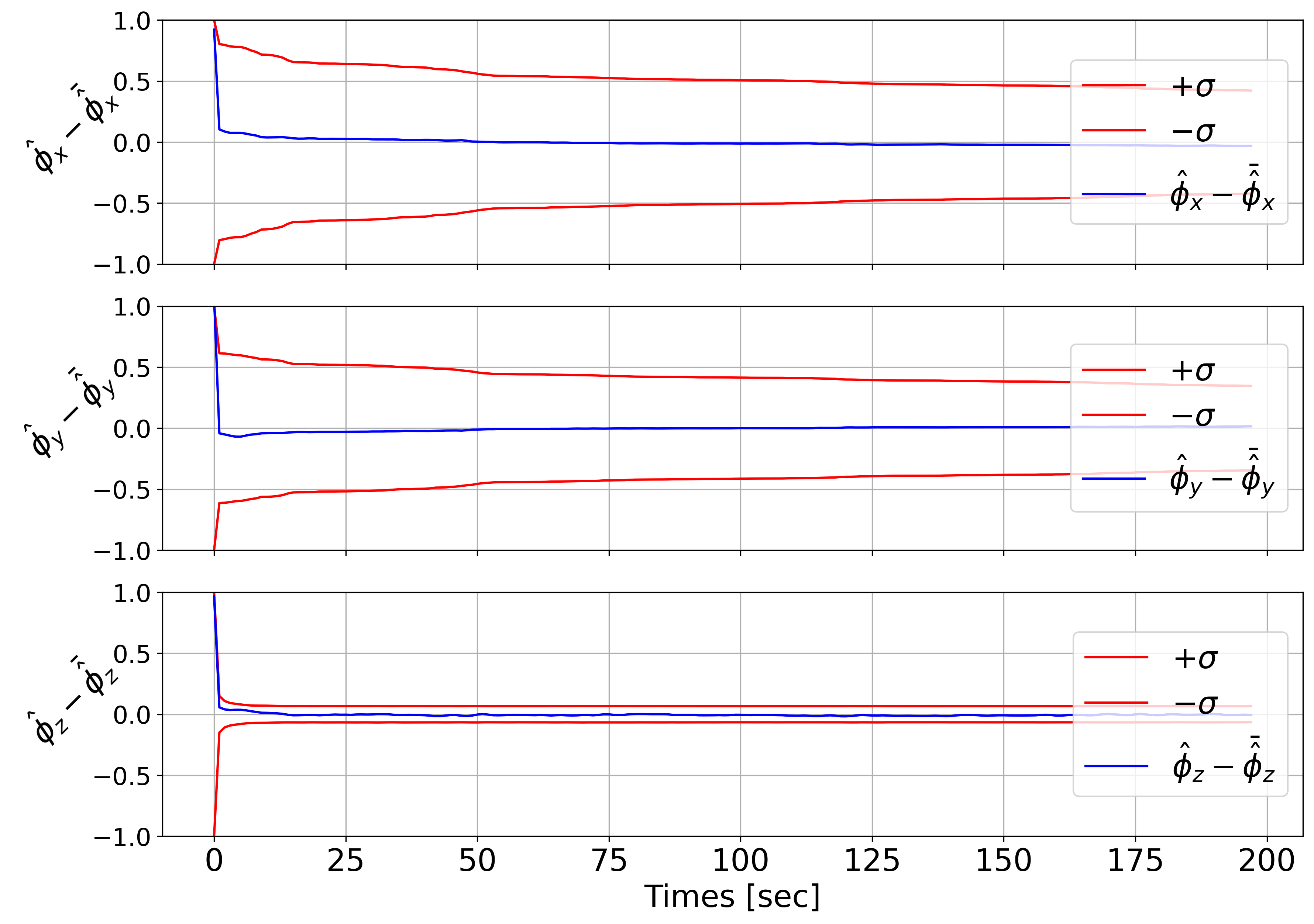}
        \label{fig:m6_cov}
    }
    \hfill
    \subfloat[Model M7]{%
        \includegraphics[width=0.32\textwidth]{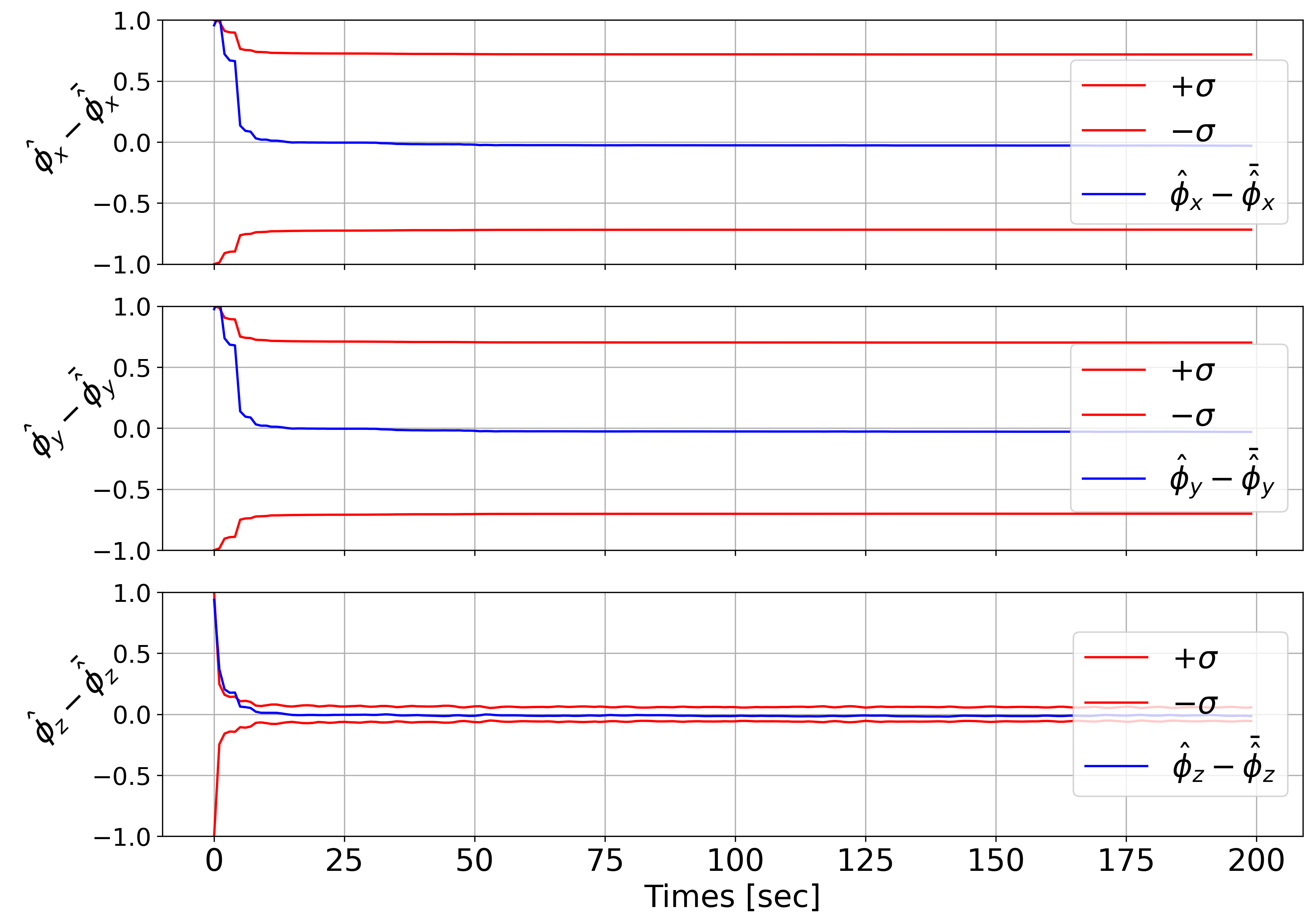}
        \label{fig:m7_cov}
    }
    \caption{State estimation of models M5 (left), M6 (middle), and M7 (right) during calibration using trajectory C1. In all three figures, the blue curves represent the state estimation error, computed as the difference between the estimated state and its average value, while the red curves represent the square root of the estimated state error covariance, corresponding to the covariance sleeve. In all three figures, the three rows show the estimated misalignment errors, $\phi_x$, $\phi_y$, and $\phi_z$.}
    \label{fig:gnss_gd_covariance_compare}
\end{figure*}

\begin{figure}
    \centering
    \includegraphics[width=0.9\linewidth]{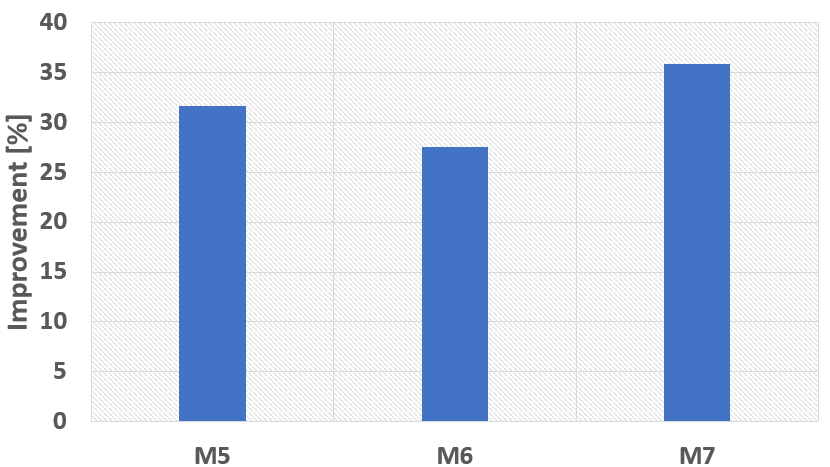}
    \caption{Proposed GNSS-free IAC approaches, M5, M6, and M7, improvements over the uncalibrated DVL baseline.}
    \label{fig:gnss_gd_improv_bar}
\end{figure}

To evaluate the IAC-GF approaches, we analyzed the state estimation error and the square root of the error state covariance sleeve. Figure \ref{fig:gnss_gd_covariance_compare} presents the state error and covariance sleeve for the sample calibration trajectory C1 from evaluation set E1.
It shows that the state estimation error remains bounded within one standard deviation of the covariance sleeve. However, M7 exhibits a slight bias, which is expected since the calibration is performed without external aiding. In the GNSS-free calibration scenario, the uncalibrated solution serves as the baseline. Therefore, for the uncalibrated case, the evaluation procedure is executed only once, with Steps 2 and 3 omitted. Consequently, the uncalibrated velocities of the fourteen evaluation trajectories are directly compared with the GT velocities in Step 5. To further evaluate the DVL self-calibration accuracy, the validation procedure was repeated using the maximum available calibration duration for each calibration trajectory, that is, the full available calibration segment. Table \ref{tbl:gnss_GD_res} presents the Avg.VRMSE results obtained using \eqref{eq:avg_vrmse_epoch} for each evaluation set, together with the overall average.
\begin{table}[h!]
\centering
\caption{Average VRMSE values, in $[\mathrm{m/s}]$, of the proposed IAC in GNSS-denied environments models comparing the uncalibrated velocity as baseline.}\label{tbl:gnss_GD_res}
\resizebox{\columnwidth}{!}{%
\begin{tabular}{|c|c|c|c|c|}
\hline
   \backslashbox{Eval. set \\ (Calib. traj)}{Model}     & Uncalibrated & M5 (Ours)       & M6 (Ours)       & M7 (Ours)       \\ \hline
E1 (C1) $[m/s]$     & 0.1185       & 0.0665          & 0.0646         & 0.0625          \\ \hline
E2 (C2) $[m/s]$    & 0.1173       & 0.0893          & 0.0945          & 0.0739          \\ \hline
E3  (C3)$[m/s]$   & 0.1197       & 0.0634          & 0.0625         & 0.0798          \\ \hline
E4  (C4)$[m/s]$   & 0.1197       & 0.1058          & 0.1229          & 0.0888          \\ \hline
Average $[m/s]$ & 0.1188       & \textbf{0.0813} & \textbf{0.0861} & \textbf{0.0762} \\ \hline
\end{tabular}%
}
\end{table}\\
Table \ref{tbl:gnss_GD_res} shows that all three proposed IAC-GF approaches outperform the uncalibrated baseline. Moreover, when C1 and C3 are used as the calibration trajectories in evaluation sets E1 and E3, respectively, models M5 and M6 achieve nearly half the Avg.VRMSE of the uncalibrated baseline. However, when C4 is used as the calibration trajectory in E4, the improvement achieved by M5 and M6 is less pronounced. These observations indicate that the proposed heuristic for approximating the sway and heave components, based on the initial two seconds of data, provides a reasonable estimate of the missing velocity components. Nevertheless, M7 delivers consistent performance across all evaluation sets, despite estimating all three states using only the surge (X-axis) measurement.\\
To further assess the effectiveness of the IAC-GF approaches, Figure \ref{fig:gnss_gd_improv_bar} presents the average Avg.VRMSE improvement relative to the uncalibrated baseline, computed from the results reported in Table \ref{tbl:gnss_GD_res}. It shows that M7 achieves the largest improvement, approximately $35\%$, followed by M5 with about $31\%$ and M6 with approximately $27\%$.
Overall, the proposed IAC-GF approaches achieve an average improvement of approximately $30\%$, despite the absence of GNSS reference measurements. These results highlight two key outcomes of the proposed IAC-GF framework. First, the proposed IAC-GF assumptions are validated, demonstrating their applicability and robustness. Second, M7 enables GNSS-free DVL self-calibration using only the velocity norm and the simplified error model, highlighting the practicality and simplicity of the proposed approach. Together, Table \ref{tbl:gnss_GD_res} and Figure \ref{fig:gnss_gd_improv_bar} demonstrate that the proposed self-calibration approaches enable DVL calibration in GNSS-denied environments and provide a significantly better alternative to uncalibrated DVL measurements, which would otherwise lead to rapid navigation drift.
\section{Conclusions}\label{sec:conc_sec}
Despite the importance of accurate DVL calibration, traditional Kalman filter-based calibration approaches neither utilize information aiding to improve accuracy nor enable calibration when GNSS signals are unavailable. To address these limitations, this work proposed two IAC frameworks for both GNSS scenarios. In GNSS-enabled environments, the core novelty of our proposed IAC-G framework relies on approximating the Z component of the GNSS velocity expressed in the DVL frame as zero to improve calibration accuracy. In GNSS-denied environments, our proposed IAC-GF framework tackles DVL calibration by combining a simplified error model, an assumption on the sway and heave components, and an approximation of the surge component to construct an a reference velocity for calibration, thereby enabling GNSS-free DVL self-calibration. Nevertheless, the proposed approaches rely on assumptions regarding the DVL Z-axis installation and approximations made as part of the IAC-GF framework for reconstructing the missing velocity components, which may be difficult to satisfy in practice.\\
Despite these limitations, our proposed IAC approaches improve calibration accuracy in both GNSS scenarios. In GNSS-enabled environments, two proposed models, M3 and M4, achieve up to a $20\%$ improvement over the best Kalman filter baseline. In GNSS-denied environments, the proposed IAC-GF approaches result in three novel models that achieve an average improvement of $30\%$ and up to $34\%$ compared to the uncalibrated baseline. Moreover, model M7 does not require prior knowledge yet achieves an impressive $35\%$ improvement, enabling fully standalone GNSS-free DVL self-calibration. All proposed approaches were validated using two separate real-world sea experiments, demonstrating their robustness and applicability.\\
These results demonstrate that the proposed IAC frameworks enable more accurate and efficient DVL calibration in both GNSS scenarios, particularly the proposed IAC-GF framework, which enables DVL self-calibration in GNSS-denied environments. The immediate outcome of the proposed IAC frameworks is improved navigation accuracy and reduced drift, which subsequently support higher-quality data collection and improved mission reliability. More broadly, this work introduces a novel information-aided paradigm for underwater navigation, opening new research directions in low-cost sensor integration, information-aided navigation pipelines, and enhancing marine research capabilities and data validity.
\section*{Acknowledgment}
\noindent Z.Y. is supported by the Maurice Hatter Foundation and University of Haifa presidential scholarship for outstanding students on a direct Ph.D. track.

\bibliographystyle{IEEEtran}
\bibliography{bio.bib}

\end{document}